# MODULAR TRAFFIC SIGNS RECOGNITION APPLIED TO ON-VEHICLE REAL-TIME VISUAL DETECTION OF AMERICAN AND EUROPEAN SPEED LIMIT SIGNS

**Fabien Moutarde and Alexandre Bargeton**

**Robotics Laboratory**
**Ecole des Mines de Paris (ParisTech)**
**60 Bd Saint-Michel, F-75006 PARIS, FRANCE**
**Tel.: (33) 1-40.51.92.92, Fax: (33) 1.43.26.10.51**

**Fabien.Moutarde@ensmp.fr**

Anne Herbin and Lowik Chanussot

Valeo Driving Assistance Domain
34 rue St-André, ZI des Vignes
F-93012 BOBIGNY, FRANCE
Tel.: (33) 1-49.42.61.63

Lowik.Chanussot@valeo.com
Benoist.Fleury@valeo.com

**Abstract:** We present a new modular traffic signs recognition system, successfully applied to both American and European speed limit signs. Our sign detection step is based only on shape-detection (rectangles or circles). This enables it to work on *grayscale* images, contrary to most European competitors, which eases robustness to illumination conditions (notably night operation). Speed sign candidates are classified (or rejected) by segmenting potential digits inside them (which is rather original and has several advantages), and then applying a neural digit recognition. The global detection rate is ~90% for both (standard) U.S. and E.U. speed signs, with a misclassification rate <1%, and no validated false alarm in >150 minutes of video. The system processes in real-time ~20 frames/s on a standard high-end laptop.

## INTRODUCTION AND RELATED WORKS

Automatic traffic signs detection and recognition (TSR) is a key module for new driving assistance smart functions, as it is a requirement for the necessary level of traffic scene understanding. For example a robust visual real-time TSR system is a pre-requisite for developing a system for reminding the driver what is the current speed limit. Some of the traffic sign information may sometime be extracted from the GPS navigation data, but it is neither always complete nor up-to-date. Moreover, temporary speed-limits for road works, as well as variable speed-limits, are by definition not included in pre-defined digital cartographic data. Therefore a visual real-time TSR system is a mandatory complement to GPS systems for designing advanced driving assistance systems.

A TSR system usually involves two main steps: 1/ detection of potential traffic signs in the image, based on the common shape/color design of sought traffic signs; 2/ classification of the selected regions of interest (ROI) for identifying the exact type of sign, or rejecting the ROI. As noted by Bahlman et al. in [1], the majority of recently published TSR approaches make use of color information (see e.g. [2], [3] or [4]), which makes the detection step easier. In contrast with that, the TSR system presented in the present paper works on grayscale images, which puts less constraint on the required sensor, and may help meet global costs requirements. Grayscale-based detection also improves robustness for operation in dark or night condition, as noted and advocated in [5] and [6]. Finally, TSR system are generally developed for only one particular country regulation, while one of the originality of our system is to be modular enough to be easily adapted to very different traffic signs designs: we present here promising results for detection and recognition of both American (U.S.) rectangular speed limit signs, and European (E.U.) circular signs.

# MODULAR SYSTEM ARCHITECTURE

## ARCHITECTURE OVERVIEW

The system presented here is implemented using the $^{RT}$MAPS software for real-time multi-sensor applications prototyping, distributed by Intempora (http://www.intempora.com), and already adopted by many French car manufacturers for the development of on-vehicle real-time applications. We take full advantage of the intrinsic modularity offered by the graphical programming paradigm of $^{RT}$MAPS, by using separate modules for the main successive steps of TSR:

- detection of potential traffic signs
- recognition
- tracking and validation

The modularity of our system allows to easily adapt it to different types of speed-limit signs (e.g. U.S. vs E.U. speed-limit signs), or even detection of other kind of traffic sign.

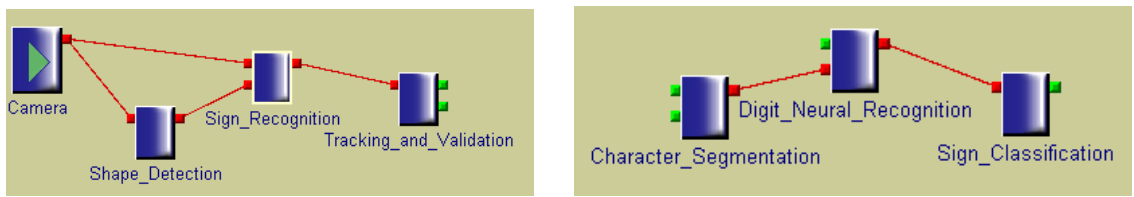

*Figure 1: Global modular architecture (left) and detail of the Sign_Recognition block for the E.U. case (right).*

## SIGN CANDIDATE DETECTION

As already mentioned we wanted our system to operate even on grayscale videos, notably for easing night-time operation. Therefore, our detection modules are based on shape-detection, as for instance in [6], but more general:

- a circular Hough-transform specially adapted and tuned for the application to European Union (E.U.) speed-limits signs, which are circular;
- a specially-designed rectangle-detection (covered by a pending patent application) based on edge detection for United States (U.S.) speed-limit signs, which are rectangular.

The aim of the detection stage is to miss as few real sought signs as possible. It is in particular essential to be able to detect efficiently even in the case of low luminosity and/or contrast of the sign contour on the background. False detections at this stage are not a problem, as they will be efficiently filtered by the recognition step, because most of the detected non-sign rectangles or circles do not even contain a single digit candidate (see detail on recognition step below).

# SIGN RECOGNITION

The current version of the recognition part itself is further subdivided in more modules, one of them trying to segment characters inside the potential speed-limit signs, and another one applying a neural-network optical digit recognition (ODR). Doing the sign recognition by extracting and recognizing digits inside the sign is one of the originality of our approach (to our knowledge, only [4] have proposed something in the same spirit, while most currently published works or developed systems for speed-limit sign recognition have chosen to do a global recognition of the whole signs). This choice was primarily motivated by the great variability of the exact text content (and even shape/size) of U.S. speed-limit signs, as shown on figure 2.

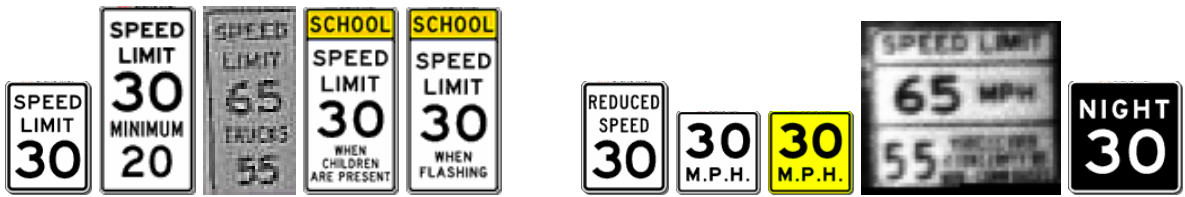

*Figure 2. Illustration of some of the many types of U.S. speed-limit signs, which makes it complicated for global sign recognition approaches.*

This design choice (using digit extraction and recognition instead of global sign recognition) proves to be an advantage also in Europe, where there are actually some differences in global aspect of the same sign among different countries, and even inside a single country such as France for instance (see figure 3). Even though we presently train a specific neural network for speed sign digits recognition (see below), a final pan-European system could probably work with a "universal" digit recognition module, therefore not requiring the previous collect of all European variants of each sign.

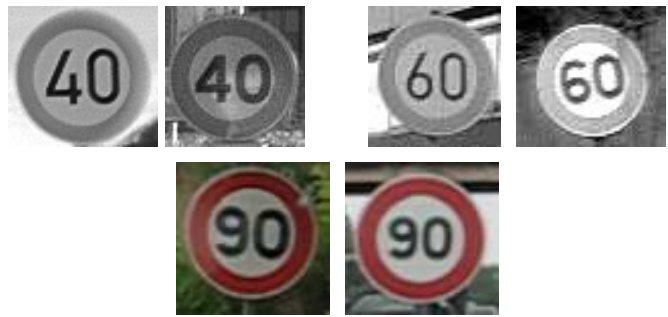

*Figure 3. Illustration of the variability of global aspect of the same speed-limit sign digits, both across different E.U. countries and even inside a given country: on top German signs (left of each pair) compared to their French equivalent, and on bottom two variants of the French 90 speed-limit sign.*

Naturally, as our system recognizes the sign based on the digits inside, we must have an efficient and robust character segmentation algorithm. We also needed it to be intrinsically insensitive to orientation variations, because signs are sometimes not perfectly vertically set, but rather slightly tilted. The segmentation used is a connected-component labeling applied after a binarization obtained by adaptive thresholding on the circular or rectangular area, as illustrated on figure 4.

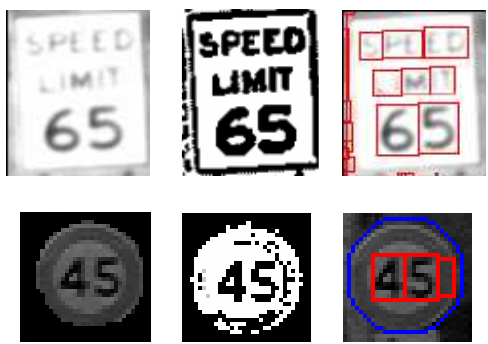

*Figure 4. Illustration of the character segmentation procedure in a rectangle (U.S. case, top line), and in a circle (E.U. case, bottom line): the searched area (left) is first binarized by adaptive thresholding (middle), then connected-component labeling is applied for finding potential digits (right).*

The neural network ODR module is a multi-layer perceptron (MLP) with 10 output (1 for each digit value) trained on specially built databases of digits extracted from relevant speed-limit signs in videos recorded from on-vehicle camera. For example, our current E.U. digits database, constantly enriched with new examples extracted from video recorded in different European countries, currently contains a total of 2762 digits examples and 2789 negative (non-digits) examples.

Note that for the U.S. case, another MLP classifier is also used to analyze the region above the digits in order to verify that the sign is actually a standard speed-limit, and not for instance a "truck speed" sign.

Depending on all the outputs of the neural network ODR module, a confidence measure is computed and assigned to the detected and recognized speed-limit signs. This confidence is further increased if the same sign is again detected and identified at nearly the same image location in subsequent video frames. The recognized speed-limit sign is finally validated if its confidence gets over a validation threshold, typically determined so that validation occurs if the sign is identified with reasonable confidence on at least 2 or 3 nearly successive frames.

## EXPERIMENTS AND RESULTS

We evaluate only the global system performance, by using video recordings independent from those used for extracting digits for training our neural ODR module, and counting the percentage of signs correctly detected and validated before they are passed. Preliminary evaluation of the U.S system showed a global system correct detection rate (SCDR) of ~90%.

The system currently recognizes the most frequent U.S. speed limit signs, i.e. those for which "SPEED LIMIT" is written just above the speed limit value (see examples on left of figure 2). This is illustrated on figures 4 and 5 which show correct detection and recognition of two different flavors of these kinds of speed-limit signs.

Other kinds of U.S. speed-limit signs such as those on right of figure 2 are not recognized by the current version, but thanks to our modular and easily parameterized architecture, it should be rather easy to modify it in order to take into account different variants such as the one where "REDUCED SPEED" is written instead of "SPEED LIMIT", or the small ones with only "M.P.H." written below.

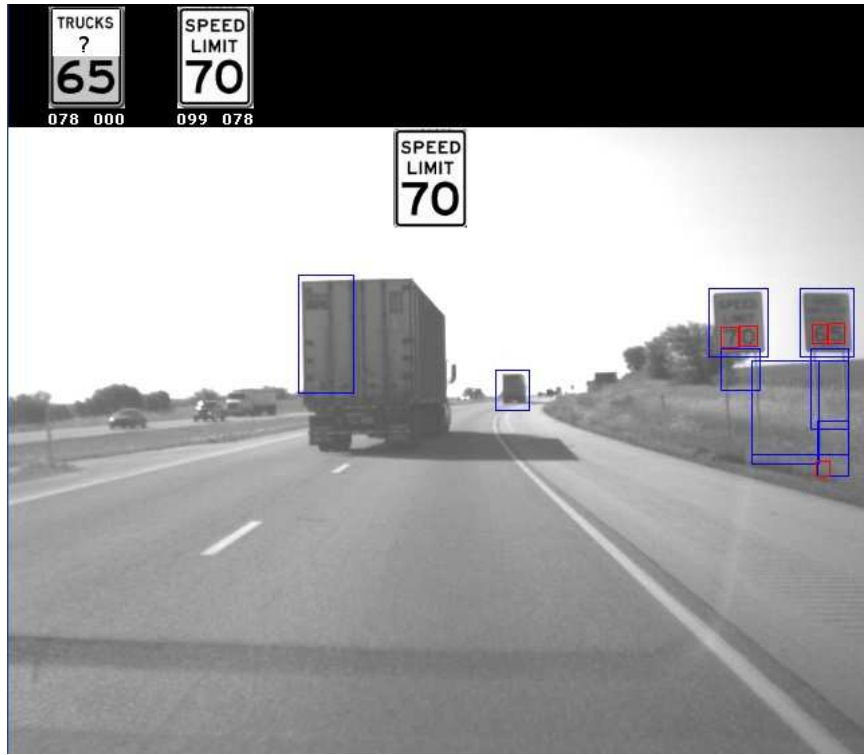

*Figure 5: Speed limit sign detection on U.S. road (case of most common type of sign)*

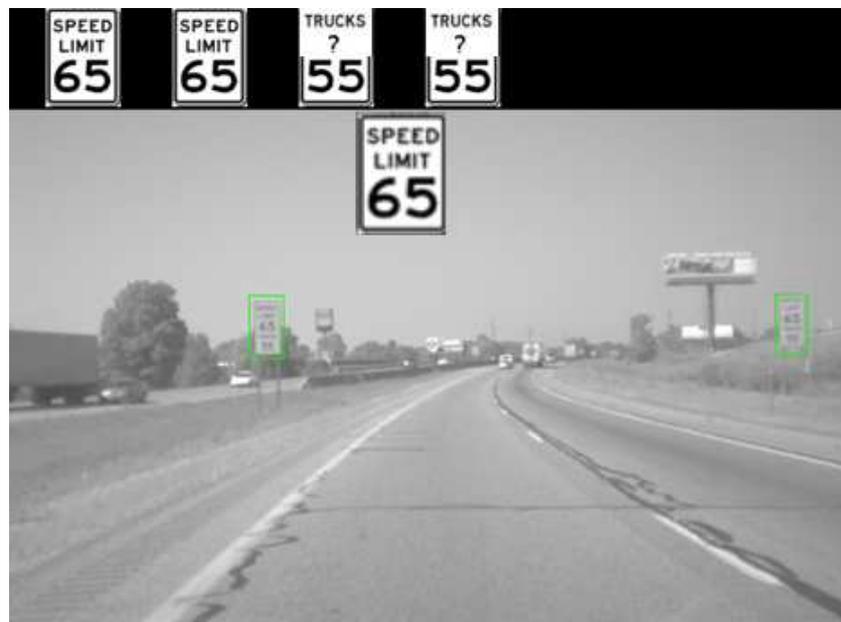

*Figure 6: Detection and recognition of another variant of U.S. speed limit sign, with exactly the same system.*

A more thorough evaluation has been done for the E.U. system, using ~150 minutes of recordings on French roads and streets, under various daytime illumination conditions, and containing 281 speed limit signs covering 12 different limit values (10, 20, 30, 40, 45, 50, 60, 70, 80, 90, 110, 130). The global system correct detection rate (SCDR) is ~89%, as shown on table 1.

| Total number of speed signs | Signs detected and validated with correct type | Missed signs (not validated) | Signs detected and validated, but misclassified |
|---|---|---|---|
| 281 | 250 | 29 | 2 |

*Table 1. First global quantitative evaluation of European speed limit sign detection (conducted on French videos).*

Most of the 11% non-correct-detections are just missed signs (sometimes because of not contrasted enough edges of the sign, and most of the time because of noise or occlusion impeding digit segmentation). The misclassification rate (signs for which a wrong speed value has been validated) is below 1%. And, most importantly, not a single *validated* false alarm has been noticed in the 150 minutes of daytime recording: all spurious signs are efficiently filtered by our tracking and validation module. Note that these 150 minutes of analyzed recording are not consecutive, but selected around moments when speed limit signs are visible, and cover various illumination conditions and types of roads and environment. Also, several experimental real-time on-road tests aboard an experimental car have been conducted, for a total of many driving hours, and were quite satisfactory.

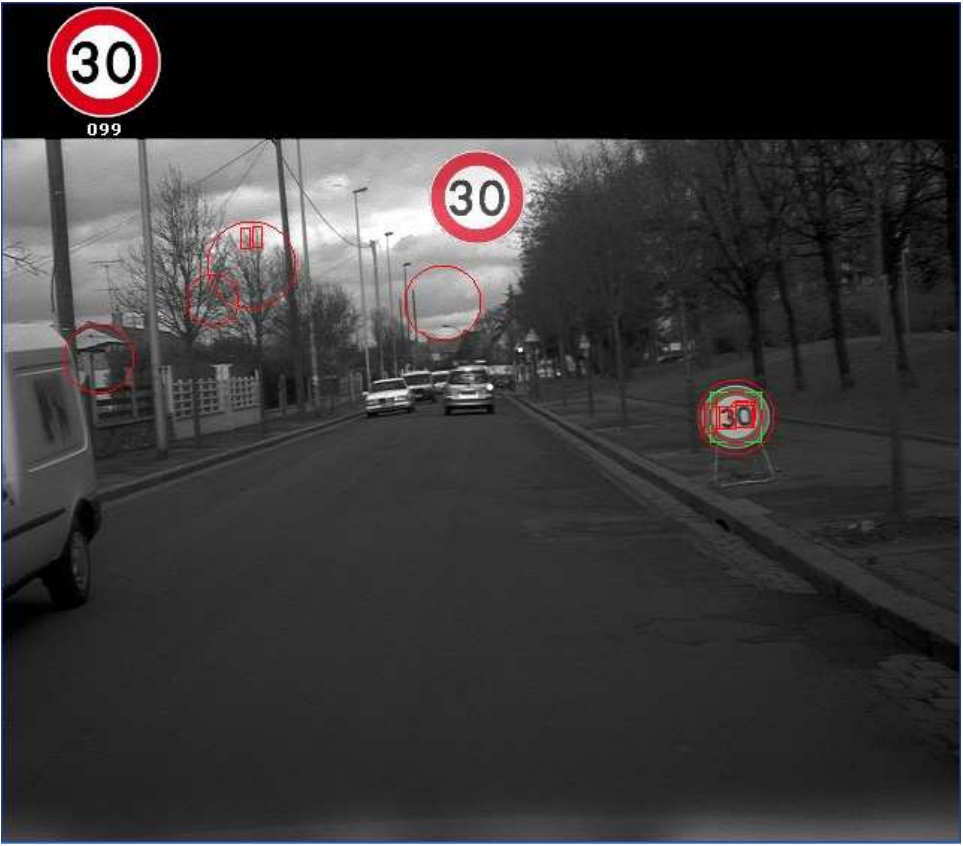

*Figure 7: Speed limit sign detection on E.U. street;*
*this example also illustrates the interest of visual detection the case of temporary roadwork*

The above results were quantified only on French videos, but evaluations in other E.U. countries are currently under way, with very promising results (see figure 8 with an example in Germany).

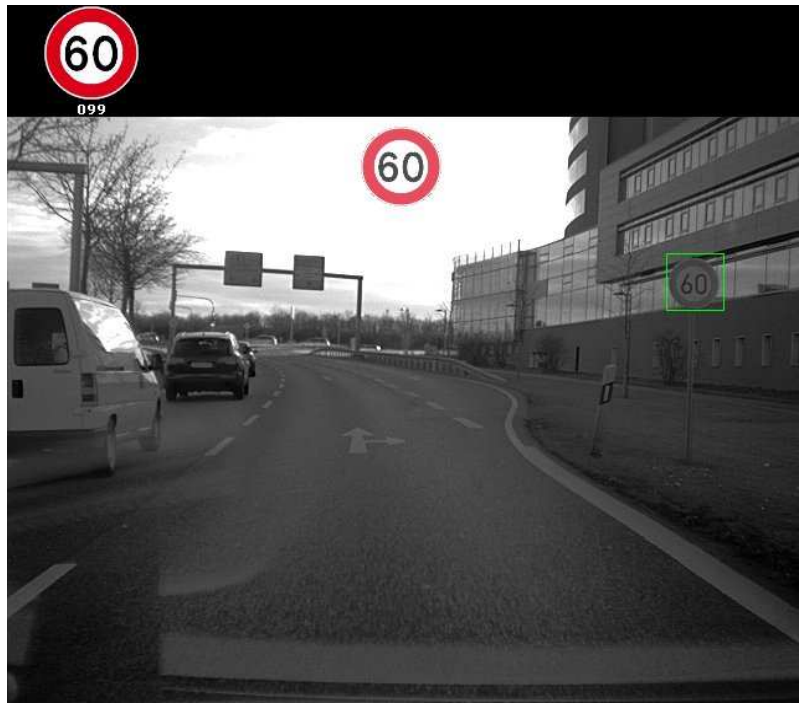

*Figure 8. Correct recognition of a German speed-limit sign illustrating promising results for pan-European speed-limit recognition.*

## CONCLUSIONS AND PERPECTIVES

We have presented a robust and effective visual speed-limit signs detection and recognition system, with 2 variants of the same global architecture working respectively for U.S. and E.U. signs, both with ~90% global correct sign detection rate. The system requires only grayscale videos, and is able to process 640x480 videos, at ~20Hz in real-time on a standard 2.13GHz dual core laptop. It has a remarkably low false alarm rate (less than 1 spurious sign in 150 minutes of operation).

Evaluation of the performances of our system at night and in tunnels is currently in progress, and encouraging. Some still ongoing work is focusing on eliminating the remaining ~1% misclassifications, and lowering the current ~10% miss-rate, by adding a complementary sign recognition scheme.

Also, a parallel work already done in another context [7] for extracting cartographic information from GPS navigation maps, has been extended for extracting also speed limit information. We have thus begun to develop a framework for fusion of the output of visually-detected speed limits with GPS cartographic speed limit data. Preliminary experiments show quite promising results for a final system that could take into account those two complementary sources. Such a system would provide accurate speed limit information even when a sign was visually occulted by another vehicle, and inversely take into account detection of temporary (e.g. roadwork-related) speed limit not included in cartographic information.